\documentclass[letterpaper, 10 pt, conference]{ieeeconf}  

\IEEEoverridecommandlockouts                              

\usepackage{graphics} 
\usepackage{epsfig} 
\usepackage{mathptmx} 
\usepackage{times} 
\usepackage{amsmath} 
\usepackage{amssymb}  
\usepackage{cite}
\usepackage{mathrsfs}
\usepackage{color}
\usepackage{booktabs}
\usepackage{makecell}
\usepackage{multirow}
\usepackage{algorithm}
\usepackage{algorithmicx}
\usepackage{diagbox}
\usepackage[noend]{algpseudocode}
\usepackage{graphicx}
\usepackage{threeparttable}
\usepackage[T1]{fontenc}
\usepackage{hyperref}
\hypersetup{hidelinks}

\usepackage[table]{xcolor}

\newcommand{\ci}[1]{\tiny{\textcolor{gray}{~($\pm #1$)}}}

\title{\LARGE \bf
Multi-objective Cross-task Learning via Goal-conditioned \\ GPT-based Decision Transformers for Surgical Robot Task Automation
}

\author{Jiawei Fu, Yonghao Long, Kai Chen, Wang Wei, Qi Dou 
\thanks{J. Fu, Y. Long, K. Chen, W. Wei, and Q. Dou are with the Department of Computer Science and Engineering, The Chinese University of Hong Kong.
This work was supported in part by a grant from the Research Grants Council of the Hong Kong Special Administrative Region, China (Project No. 24209223), in part by the Hong Kong Innovation and Technology Fund (Project No. ITS/223/22), in part by a grant from National Natural Science Foundation of China (Project No. 62322318), and in part by InnoHK Multi-scale Medical Robotics Centre.}
\thanks{Corresponding author: Qi Dou (qidou@cuhk.edu.hk)
}}

\begin{document}

\maketitle
\thispagestyle{empty}
\pagestyle{empty}

\begin{abstract}

Surgical robot task automation has been a promising research topic for improving surgical efficiency and quality. Learning-based methods have been recognized as an interesting paradigm and been increasingly investigated. However, existing approaches encounter difficulties in long-horizon goal-conditioned tasks due to the intricate compositional structure, which requires decision-making for a sequence of sub-steps and understanding of inherent dynamics of goal-reaching tasks. In this paper, we propose a new learning-based framework by leveraging the strong reasoning capability of the GPT-based architecture to automate surgical robotic tasks. The key to our approach is developing a goal-conditioned decision transformer to achieve sequential representations with goal-aware future indicators in order to enhance temporal reasoning. Moreover, considering to exploit a general understanding of dynamics inherent in manipulations, thus making the model's reasoning ability to be task-agnostic, we also design a cross-task pretraining paradigm that uses multiple training objectives associated with data from diverse tasks. We have conducted extensive experiments on 10 tasks using the surgical robot learning simulator SurRoL~\cite{long2023human}. The results show that our new approach achieves promising performance and task versatility compared to existing methods. The learned trajectories can be deployed on the da Vinci Research Kit (dVRK) for validating its practicality in real surgical robot settings. Our project website is at: \url{https://med-air.github.io/SurRoL}.

\end{abstract}

\vspace{-0.14cm}
\section{Introduction}

Surgical robot task automation has been increasingly studied for its potential to improve surgical efficiency and augment robot intelligence. Recent advancements have witnessed research on learning-based methods~\cite{long2023human,huang2023guided,seita2020deep,su2020reinforcement,chi2020collaborative} to promote automation of surgical robots. Still, current performances of the latest methods are impeded in long-horizon goal-conditioned tasks, where a sequence of actions and sub-steps are required until reaching an ultimate goal. Previous algorithms with reinforcement learning~\cite{sutton2018reinforcement} and Markov decision process only predict actions from the current state while overlooking information from historical sequential states and actions. This lacks temporal reasoning capability over actions and affects learning of the inherent sequential dynamics which is useful to the final success of a complex task. Despite some works~\cite{huang2023value, lee2021adversarial} combining task-specific strategies to conduct sub-task decomposition, those customized designs may sacrifice the generalization ability to other surgical robot tasks, thus imposing constraints on their overall efficacy and applicability in the domain.

In this regard, to address long-horizon goal-conditioned tasks in surgical robot learning, the model should understand the contextual dependency of the sequence across diverse tasks, and hold holistic features of the goal-reaching pattern in tasks to accurately predict appropriate decisions. Recently, large language models (LLMs) are very popular through the success of GPT~\cite{radford2018improving}, and have been transferred to solving decision-making problems in robotics~\cite{NEURIPS2021_7f489f64}. Researchers apply the transformer backbone of LLMs for decision-making from historical state and sequential action input. Specifically, decision transformer families~\cite{NEURIPS2021_7f489f64,pmlr-v202-yamagata23a,pmlr-v162-zheng22c,lee2022multi,hu2023decision} have achieved impressive results in gym- and atari-based environments. They use autoregressive models to yield the trajectories through the reasoning of transformer architectures, and forecast the action based on the history sequence and returns-to-go, which means an accumulated reward from a current state to the ultimate goal state without decay.

Nevertheless, existing decision transformer methods~\cite{NEURIPS2021_7f489f64,pmlr-v202-yamagata23a,pmlr-v162-zheng22c,lee2022multi,hu2023decision} heavily rely on immediate reward feedback to update returns-to-go, which cannot be satisfied in our considered goal-conditioned paradigm, because the agent would only get rewards when it reaches final goals in surgical tasks. Furthermore, the introduction of task-specific rewards and the loss of cross-task pretraining create varying internal dynamics across tasks, resulting in technical challenges in developing a unified framework for reasoning and decision-making within the goal-reaching paradigm in surgical tasks.

To leverage the advanced GPT-based decision-making frameworks for improving surgical robot task automation, we propose the goal-conditioned decision transformer that embedds goal and time-to-goal as future indicators. 
Besides, we formulate multiple training objectives: \textit{action prediction}, \textit{dynamics prediction}, \textit{time-to-goal prediction}, and \textit{sequence reconstruction} in our cross-task pretraining process, which fosters a comprehensive representation of the temporal dynamics inherent in goal-conditioned tasks and encourages the model to incorporate diverse temporal reasoning factors.
Based on such pretraining, the model is updated on the targeted specific task as a downstream use case. Importantly, the update process does not require any modification on the model architectures, thanks to the shared goal-reaching pattern which is uniform regardless of the individual reward of each task. 
We rely on the open-source simulator SurRoL~\cite{xu2021surrol,long2023human} to collect data and conduct model training. Experimental results show that our proposed method exceeds existing decision-making algorithms and task-specific methods in average performance and versatility among diverse tasks.
The contributions are:
\begin{itemize}
    \item We propose a new learning-based framework with goal-conditioned decision transformer for surgical robot scenarios, which is designed for goal-reaching surgical tasks with a strong reasoning ability. 
    \item We design a multi-objective cross-task pretraining strategy, which learns the sequential context dependency inherent in data from diverse tasks to reason the internal dynamics of goal-conditioned paradigms.
    \item Experimental results demonstrate superior performance of our proposed approach compared with state-of-the-art learning-based methods. Moreover, we deploy the trajectory of our method into real-world dVRK platform to show its practicality.
\end{itemize}

\section{Related Work}

\subsection{Decision-making with transformer architectures}

Large-scale transformers have propelled LLMs to achieve breakthroughs in natural language processing tasks~\cite{brown2020language, ouyang2022training} with the context learning capacity. Inspired by its long-term trajectory modeling, decision transformer~\cite{NEURIPS2021_7f489f64} first utilizes transformer architecture for sequential decision-making problems, where sequences constructed from history states, actions and returns-to-go are embedded in transformers to replace the previously used text information. It applies autoregressive models for the sequence and can incorporate the instantaneous task-dependent reward function to allow the model be aware of future states. Yamagata \textit{et al.}~\cite{pmlr-v202-yamagata23a} learn the $Q$ value additionally to help find the optimal action trajectory based on the decision transformer. Online decision transformer~\cite{pmlr-v162-zheng22c} modifies to stochastic policy and applies extra interactions with environments, which improves its performance. Hu \textit{et al.}~\cite{hu2023decision} augment the decision transformer architecture to accommodate tasks characterized by missing frames through the implementation of random masking techniques, alongside the exclusion of temporal span encoding. 


This approach is designed to investigate the transformer's capability to decipher the sequence of events and the intricacies involved in the dynamics of the goal-conditioned paradigm. By employing a multi-task pretraining scheme across different tasks, we aim to cultivate a better understanding of the underlying dynamics representative to goal-conditioned tasks, and further improve overall performance.

\subsection{Surgical task automation}

In recent years, surgical robot automation has emerged as a popular area of research. Existing works have designed rule-based methods to improve the performance on multiple tasks, such as endoscope manipulation~\cite{osa2010framework, king2013towards, rivas2019transferring}, surgical suturing~\cite{schwaner2021autonomous, leonard2014smart}, tissue cutting~\cite{khadem2016mechanics,patel2018using}, etc. However, the rule-based methods require sophisticated design for every possible case, which consumes labor and lacks generalization probability to new tasks. To overcome these weaknesses, learning-based algorithms, i.e., reinforcement learning and imitation learning are increasingly studied in surgical robotics~\cite{huang2023guided,seita2020deep,su2020reinforcement,chi2020collaborative,ben2023learning}. Huang \textit{et al.}~\cite{huang2023guided} embed expert demonstrations in reinforcement learning to tackle the large exploration burden in learning for completing surgical tasks. To improve the performance of long-horizon tasks, sophisticated task-dependant sub-goals are designed~\cite{huang2023value} for training a chaining policy to connect the sequence of actions. In addition, Seita \textit{et al.}~\cite{seita2020deep} apply deep imitation learning to achieve sequential fabric smoothing in dVRK platform from raw sensor inputs.

Previous learning-based methods~\cite{huang2023value,lee2021adversarial} address long sequence tasks with complex task decomposition, which only adapts to a few tasks that are carefully designed and modified. This pattern results in poor generalization performance and needs a significant amount of manually designed work. To alleviate this problem, we utilize the reasoning and good generalization~\cite{lee2022multi} ability of transformers as demonstrated in LLMs for surgical task automation. Our model can be versatile and general without extra task-specific designs.

\section{Method}

We first provide our target problem in Sec. \ref{problem_formulation_subsection}. We then present the details of our proposed framework in Sec. \ref{model_architecture_subsection}, which contains trajectory representation, architecture, and the training and evaluation schemes. Next, we show the multi-objective pretraining and downstream task learning in Sec. \ref{training_details_subsection}. Finally, the Sec. \ref{data_augmentation_and_implementation_details_subsection} describes the offline hindsight data augmentation strategy and our implementation details.  
\begin{figure*}[t]
    \centering
    \includegraphics[width=1.6\columnwidth]{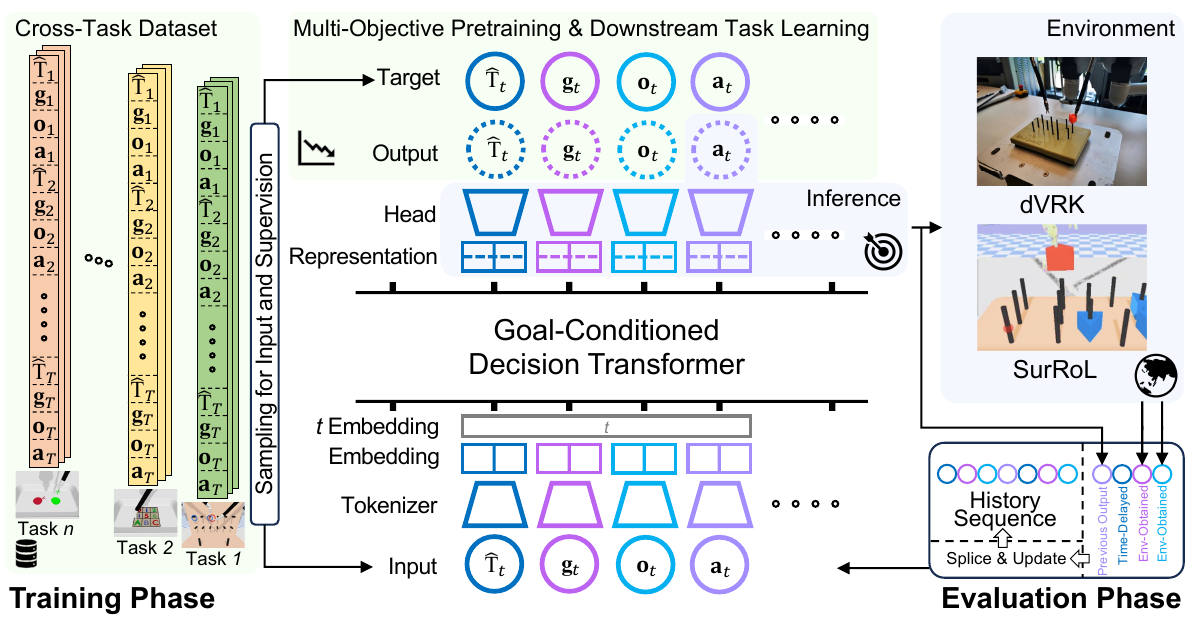}
    \setlength{\abovecaptionskip}{-0.15cm}
    \caption{Illustration of the architecture of the proposed model. For each timestep $t$, the sequence consists of four items: ${\hat{T}}_{t}$ (time-to-goal), $\mathbf{g}_{t}$ (goal), $\mathbf{o}_{t}$ (observation), $\mathbf{a}_{t}$ (action), which are embedded with the embedding of timestep and processed by the GPT architecture transformer backbone. In summary, the GPT backbone processes the input to predict results via specific heads. During pretraining and learning, sequences act as both input and target, guided by training objectives. In evaluation, we update a cached history sequence with model predictions and environmental data to forecast action $\mathbf{a}_{t}$.}
    \label{network_architecture_fig}
    \vspace{-6mm}
\end{figure*}

\subsection{Problem formulation}

\label{problem_formulation_subsection}

Goal-conditioned surgical automation tasks model those tasks specifying a goal $\bf{g}$ for the agent to achieve. The $\bf{g}$ can be described by the final states of the surgical robot arms or the manipulated objects, such as desired poses or orientations. Combining the time-ordered observation sequence $\{{\bf{o}}_{t'}\}_{t'=1}^{t}$ and the history executed action sequence $\{{\bf{a}}_{t'}\}_{t'=1}^{t-1}$, our target is to leverage transformer to represent and optimize a policy $\pi({\bf{a}}_{t}|\{{\bf{o}}_{t'}\}_{t'=1}^{t}, \{{\bf{a}}_{t'}\}_{t'=1}^{t-1}, \bf{g})$ that produces an action ${\bf{a}}_{t}$ for current execution and maximize the probability that the agent will achieve $\bf{g}$. We note that the goal-reaching pattern is different from the conventional reward-maximizing formulation, where environments set task-specific reward $r$. The previous transformer-based decision-making methods are designed to construct a policy $\pi({\bf{a}}_{t}|\{{\bf{o}}_{t'}\}_{t'=1}^{t}, \{{\bf{a}}_{t'}\}_{t'=1}^{t-1})$ with the help of returns-to-go ${\hat{R}}_{t}:=\sum_{t'=t}^{T} r_{t'}$, where $T$ is total steps to the ultimate goal and $t$ is current timestep. It guides the agent to achieve more accumulative rewards. 

\subsection{Policy learning via goal-conditioned decision transformer}

\label{model_architecture_subsection}

Trajectory representation is significant for our method. The selected representation should involve the useful temporal feature and can guide the model to reason for the future. In this regard, we additionally embed the goal $\bf{g}$ in the representation. Besides, different from previous methods~\cite{NEURIPS2021_7f489f64,pmlr-v162-zheng22c,hu2023decision,lee2022multi,pmlr-v202-yamagata23a} using returns-to-go, which depends on the instantaneous feedback rewards from environments, we select time-to-goal $\hat{T}$, which means the number of timesteps from the current state to the final goal in trajectory. The reason for this design comes from that goal-conditioned tasks will not give any rewards until the final targets, the accumulated future reward of returns-to-go will only be zero until arriving at the goal, which lacks the information to guide the model to be future-aware. To satisfy the aforementioned conditions and improve the sequence reasoning capacity of the model, our trajectory representation consists of four items in temporal order:
\begin{align*}
    \xi := \{({\hat{T}}_{1}, \mathbf{g}_{1}, \mathbf{o}_{1}, \mathbf{a}_{1}),..., ({\hat{T}}_{t}, \mathbf{g}_{t}, \mathbf{o}_{t}, \mathbf{a}_{t}),...,({\hat{T}}_{T}, \mathbf{g}_{T}, \mathbf{o}_{T}, \mathbf{a}_{T})\},
\end{align*}
where ${\hat{T}}_{t}$, ${\mathbf{g}}_{t}$, $\mathbf{o}_{t}$, $\mathbf{a}_{t}$ denote the time-to-goal, goal, observation and action at timestep $t$, respectively. 
As depicted in Fig.~\ref{network_architecture_fig}, the items in the trajectory undergo tokenization using distinct linear layers. These tokenized representations are subsequently subjected to layer normalization. Next, the timesteps are embedded alongside their corresponding normalized representations from each tokenized element in $\xi$. This combined input is processed by the transformer backbone, employing the GPT architecture to generate the learned hidden representation for the input sequence. Ultimately, the corresponding heads would organize the output from the learned representation.

Our training process is divided into two stages, i.e., multi-objective cross-task pretraining and downstream task skill learning. Both stages aim to supervise the output decoded components from the corresponding prediction heads using the input sequence and confer sequence reasoning capacity to the model with available data. The details of the pretraining and skill learning for tasks are described in Section~\ref{training_details_subsection}. In the evaluation stage, we store the executed history sequence with the same formulation as the training phase and input the history trajectory without the current needed action to the model to get the predicted action from the action head. In the subsequent iteration, we incorporate the newly acquired items into the history sequence and input the updated trajectory to generate forecast action once again. To limit the input sequence length, we abandon the items in the beginning timestep if the size of the input trajectory exceeds the maximum limitation. Moreover, we estimate time-to-goal using a time-delayed rule from the expected overall timestep for tasks, since the loss of ground truth goal-reached timestep in the evaluation phase.




\subsection{Cross-task pretraining and downstream task learning}

\label{training_details_subsection}

\begin{figure*}[t]
    \centering
    
    \includegraphics[width=1.0\textwidth]{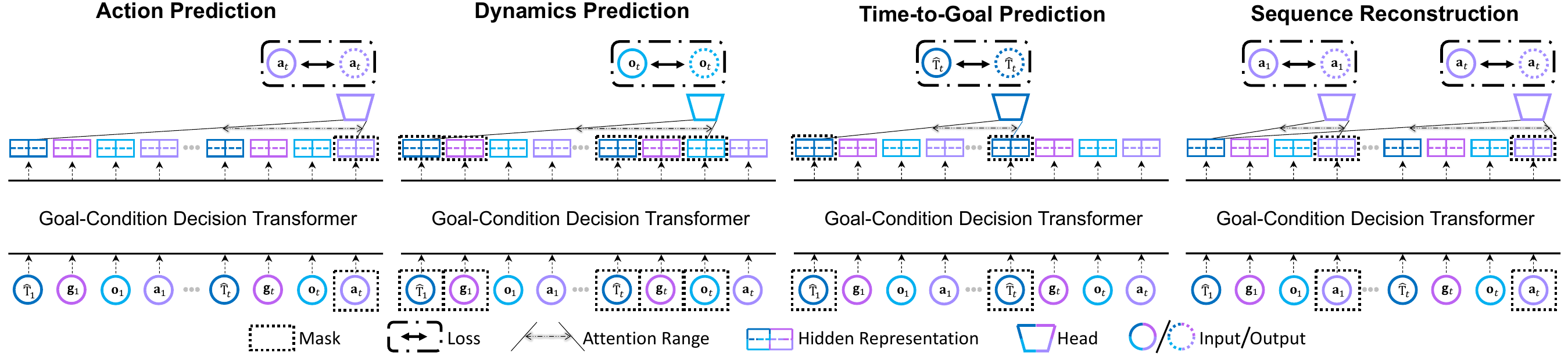}
    \setlength{\abovecaptionskip}{-0.3cm}
    \vspace{-0.3cm}
    \caption{Illustration of multiple training objectives that boost sequence contextual reasoning and understanding of task-agnostic paradigms of the model.}
    \vspace{-6mm}
    \label{multi_objective_training_fig}
\end{figure*}

\subsubsection{Cross-task multi-objective pretraining}

\label{pretraining_details_subsection}

In the pretraining stage, we first augment all individual data from several tasks using the data augmentation method. The augmented cross-task dataset is deployed to train the GPT backbone of our model with task-specific tokenizers and prediction heads. This operation aims to maximize the use of existing data to improve the temporal reasoning and generalizability of our model. Furthermore, based on the deployed GPT backbone, we have designed four training objectives with corresponding masks to offer multiple forms of supervision and enhance the overall contextual reasoning capability of our model. These objectives encompass action prediction, forward dynamics prediction, time-to-goal prediction, and sequence reconstruction. The involved objectives and their masks are depicted in Fig. \ref{multi_objective_training_fig}. The first three terms focus on both short-term and long-term relationship understanding for our model. The last term is devised to let our model study global and priori features from hindsight views. All the involved four objectives follow the goal-conditioned paradigm without tasks-specific rewards, which ensures the generalization and transferability among all the training tasks. By optimizing these objectives jointly, the model is compelled to consider a diverse set of temporal factors over sequences and understand the goal-conditioned paradigm behind all tasks to produce precise decisions.

\textit{Action prediction:} The result of action prediction directly determines the overall decision-making performance of our model. We input the recorded sequence into our model, and the action prediction head outputs the forecast action from the learned representation with the reasoning ability of GPT architecture. We compute the MSE loss between the predicted and original actions in sequence to supervise the learning process. 

\textit{Forward dynamics prediction:} Forward dynamics illustrates the internal physics property of the tasks, which plays a crucial role in boosting the final performance of our method. During this process, we manually mask all $\hat{T}$ and $\bf{g}$ in the sequence. Given timestep $t$, our model will predict ${\bf{o}}_{t}$ given $\{({\bf{o}}_{t'}, {\bf{a}}_{t'})\}_{t'=1}^{t-1}$ based on the long sequence modeling ability of GPT. This mechanism helps the model understand and analyze the influence of both short-term and long-term information on the dynamics of tasks. 

\textit{Time-to-goal prediction:} To forecast the time-to-goal ${\hat{T}}_{t}$ at timestep $t$, we mask $\hat{T}$ in the sequence to avoid injection of priori information into the model. We prepare the input sequence as $\{({\bf{g}}_{t'}, {\bf{o}}_{t'}, {\bf{a}}_{t'})\}_{t'=0}^{t-1}\}$. Then we get ${\hat{T}}_{t}$ from the time-to-goal prediction head and calculate the loss to let the model learn to reason how long it still needs to get close to the ultimate goal and complete the task.

\textit{Sequence reconstruction:} The sequence reconstruction is inspired by BERT~\cite{devlin2018bert}. Given the entire input sequence $\{({\hat{T}}_{t}, {\bf{g}}_{t}, {\bf{o}}_{t}, {\bf{a}}_{t})\}_{t=0}^{T}$, we randomly mask items and let the model rebuild them with the corresponding prediction head from the learned hidden state by the GPT backbone. This objective helps the model understand the total-sequence-level and temporal reasoning between each item and capture long-term interrelationships among the sequence items.

\subsubsection{Downstream task learning}

In the downstream task learning stage, we incorporate augmented data tailored to the specific task at hand and ignore the data from other tasks. The reason is that in the task learning stage, data from other tasks is invalid for achieving better performance in the decision-making problem for the specific downstream task. In the learning process of this period, we exclusively focus on the \textit{action prediction} objective, while disregarding other objectives. This deliberate choice stems from our intention to leverage the inherent temporal reasoning ability and comprehension of goal-conditioned dynamic patterns of the pretrained transformer backbone to optimize action performance to the fullest extent, but not boost the reasoning ability of the GPT backbone.

\subsection{Data augmentation and implementation details}

\label{data_augmentation_and_implementation_details_subsection}

\subsubsection{Data augmentation of limited demonstrations}

\label{data_augmentation_subsection}

We implement a data augmentation strategy through the application of a hindsight relabeling technique ~\cite{andrychowicz2017hindsight}. Specifically, given the original dataset denoted as $\mathcal{D}_{\text{original}}:=\{\xi^{i}\}_{i=1}^{N}$, where $N$ is the trajectory number in the dataset. We systematically navigate through $\mathcal{D}_{\text{original}}$, performing truncation on each instance $\xi^{i}$ at every timestep $t$. This process involves substituting the intended goal with the achieved goal at that timestep, resulting in a modified instance $\xi_{\text{relabel}}$. The collection of all such modified instances $\xi_{\text{relabel}}$, derived from traversing all $\xi$ across all conceivable truncated timesteps $t$, is aggregated to form the relabeled dataset $\mathcal{D}_{\text{relabel}}$. Consequently, the comprehensive training dataset is constituted by the union of the original and relabeled datasets, expressed as $\mathcal{D}_{\text{training}}:=\mathcal{D}_{\text{original}} \cup \mathcal{D}_{\text{relabel}}$. This augmentation process not only facilitates optimal utilization of the limited dataset but also embodies the principles of hindsight to densify the goal space, thereby significantly enhancing the efficacy of our proposed methodology.

\subsubsection{Implementation details}

\label{Implementation_details_subsection}

Our model is built based on the GPT backbone~\cite{radford2018improving} with 8 layers and 4 heads. All tokenizers and prediction heads are linear single layers. The maximum encoding sequence length is 100. For both the cross-task pretraining and downstream task skill learning, we supervise all training objectives using the MSE loss. We utilize the AdamW optimizer~\cite{loshchilov2018fixing} with the initial learning rate 1e-4 and weight decay 1e-4. For real-world deployment, we execute the AI-predicted trajectory from the simulator to the dVRK platform. The code, trained models and implemented details in this work are released and integrated into the policy learning engine of the SurRoL simulator in its updated repo: \url{https://github.com/med-air/SurRoL}

\section{Experiments}

\begin{table*}[t]
\setlength{\abovecaptionskip}{-0.15cm}
    \caption{Main comparison results with recent state-of-the-art methods using mean and standard deviation of success rate.}
    \centering
    \footnotesize
    \setlength{\tabcolsep}{0.85mm}
    \begin{threeparttable}

    \begin{tabular*}{2.0\columnwidth}{c|cccccccccc}
    \toprule
    \makecell[c]{Task} & \makecell[c]{\scriptsize {NeedleReach}} & \makecell[c]{\scriptsize {GauzeRetrieve}} & \makecell[c]{\scriptsize {NeedlePick}} & \makecell[c]{\scriptsize \text{PegTransfer}} & \makecell[c]{\scriptsize {NeedleRegrasp}} & \makecell[c]{\scriptsize {BiPegTransfer}} & \makecell[c]{\scriptsize {BiPegBoard}}& \makecell[c]{\scriptsize {MatchBoardPanel}}& \makecell[c]{\scriptsize {PickAndPlace}} & \makecell[c]{\scriptsize {MatchBoard}}  \\
    \makecell[c]{dim($\mathbb{O} | \mathbb{G} | \mathbb{A}$)} & \scriptsize \makecell[c]{$\mathbb{R}^{7} | \mathbb{R}^{3} | \mathbb{R}^{5}$} & \scriptsize \makecell[c]{$\mathbb{R}^{19} | \mathbb{R}^{3} | \mathbb{R}^{5}$} & \scriptsize \makecell[c]{$\mathbb{R}^{19} | \mathbb{R}^{3} | \mathbb{R}^{5}$} &\scriptsize \makecell[c]{$\mathbb{R}^{19} | \mathbb{R}^{3} | \mathbb{R}^{5}$} & \scriptsize \makecell[c]{$\mathbb{R}^{35} | \mathbb{R}^{3} | \mathbb{R}^{10}$} & \scriptsize \makecell[c]{$\mathbb{R}^{35} | \scriptsize\mathbb{R}^{3} | \mathbb{R}^{10}$} & \scriptsize \makecell[c]{$\mathbb{R}^{35} | \mathbb{R}^{3} | \mathbb{R}^{10}$} & \scriptsize $\mathbb{R}^{31} | \mathbb{R}^{6} | \mathbb{R}^{5}$ & \scriptsize $\mathbb{R}^{31} | \mathbb{R}^{6} | \mathbb{R}^{5}$ & \scriptsize $\mathbb{R}^{20} | \mathbb{R}^{3} | \mathbb{R}^{5}$ \\
    \midrule
    \makecell[c]{\textbf{PLAS}\cite{zhou2021plas}}& \makecell[c]{\textbf{1.00}\ci{.00}}& \makecell[c]{0.00\ci{.00}}& \makecell[c]{0.00\ci{.00}}& \makecell[c]{0.16\ci{.08}}& \makecell[c]{0.00\ci{.00}}& \makecell[c]{0.00\ci{.00}}& \makecell[c]{0.00\ci{.00}}& \makecell[c]{0.00\ci{.00}}& \makecell[c]{0.00\ci{.00}}& \makecell[c]{0.00\ci{.00}} \\  
    \makecell[c]{\textbf{IQL}\cite{kostrikov2021offline}}& \makecell[c]{\textbf{1.00}\ci{.00}}& \makecell[c]{0.40\ci{.00}}& \makecell[c]{0.05\ci{.04}}& \makecell[c]{0.82\ci{.20}}& \makecell[c]{0.04\ci{.01}}& \makecell[c]{0.09\ci{.05}}& \makecell[c]{0.00\ci{.00}}& \makecell[c]{0.00\ci{.00}}& \makecell[c]{0.00\ci{.00}}& \makecell[c]{0.00\ci{.00}} \\ 
    \makecell[c]{\textbf{BC}\cite{bain1995framework}}& \makecell[c]{\textbf{1.00}\ci{.00}}& \makecell[c]{0.07\ci{.05}}& \makecell[c]{0.21\ci{.06}}& \makecell[c]{0.56\ci{.11}}& \makecell[c]{0.09\ci{.03}}& \makecell[c]{0.09\ci{.05}}& \makecell[c]{0.00\ci{.00}}& \makecell[c]{0.00\ci{.00}}& \makecell[c]{0.00\ci{.00}}& \makecell[c]{0.00\ci{.00}} \\ 
    \makecell[c]{\textbf{VINN}\cite{pari2021surprising}}& \makecell[c]{0.89\ci{.06}}& \makecell[c]{0.01\ci{.02}}& \makecell[c]{0.02\ci{.02}}& \makecell[c]{0.05\ci{.04}}& \makecell[c]{0.01\ci{.02}}& \makecell[c]{0.00\ci{.00}}& \makecell[c]{0.02\ci{.05}}& \makecell[c]{0.00\ci{.00}}& \makecell[c]{0.00\ci{.00}}& \makecell[c]{0.00\ci{.00}} \\ 
    \makecell[c]{\textbf{DT}\cite{NEURIPS2021_7f489f64}}& \makecell[c]{0.95\ci{.08}}& \makecell[c]{0.68\ci{.20}}& \makecell[c]{0.89\ci{.10}}& \makecell[c]{0.77\ci{.15}}& \makecell[c]{0.66\ci{.18}}& \makecell[c]{0.15\ci{.12}}& \makecell[c]{0.67\ci{.09}}& \makecell[c]{0.29\ci{.19}}& \makecell[c]{0.18\ci{.04}}& \makecell[c]{0.21\ci{.09}} \\ 

    \makecell[c]{\textbf{DDPGBC}\cite{nair2018overcoming}}& \makecell[c]{\textbf{1.00}\ci{.00}}& \makecell[c]{0.63\ci{.11}}& \makecell[c]{0.91\ci{.05}}& \makecell[c]{0.48\ci{.22}}& \makecell[c]{0.05\ci{.08}}& \makecell[c]{0.00\ci{.00}}& \makecell[c]{0.00\ci{.00}}& \makecell[c]{0.00\ci{.00}}& \makecell[c]{0.04\ci{.01}}& \makecell[c]{0.05\ci{.03}} \\ 
    \makecell[c]{\textbf{AMP}\cite{peng2021amp}}& \makecell[c]{0.99\ci{.02}}& \makecell[c]{0.00\ci{.00}}& \makecell[c]{0.00\ci{.00}}& \makecell[c]{0.00\ci{.00}}& \makecell[c]{0.00\ci{.00}}& \makecell[c]{0.00\ci{.00}}& \makecell[c]{0.00\ci{.00}}& \makecell[c]{0.00\ci{.00}}& \makecell[c]{0.00\ci{.00}}& \makecell[c]{0.00\ci{.00}} \\ 
    \makecell[c]{\textbf{CoL}\cite{goecks2019integrating}}& \makecell[c]{\textbf{1.00}\ci{.00}}& \makecell[c]{0.71\ci{.16}}& \makecell[c]{0.96\ci{.05}}& \makecell[c]{0.58\ci{.23}}& \makecell[c]{0.04\ci{.07}}& \makecell[c]{0.01\ci{.02}}& \makecell[c]{0.00\ci{.00}}& \makecell[c]{0.00\ci{.00}}& \makecell[c]{0.02\ci{.05}}& \makecell[c]{0.04\ci{.05}} \\ 
    \makecell[c]{\textbf{AWAC}\cite{nair2020awac}}& \makecell[c]{0.94\ci{.20}}& \makecell[c]{0.43\ci{.43}}& \makecell[c]{0.26\ci{.33}}& \makecell[c]{0.31\ci{.32}}& \makecell[c]{0.00\ci{.00}}& \makecell[c]{0.00\ci{.00}}& \makecell[c]{0.00\ci{.00}}& \makecell[c]{0.00\ci{.00}}& \makecell[c]{0.00\ci{.00}}& \makecell[c]{0.07\ci{.04}} \\ 
    \makecell[c]{\textbf{DEX}\cite{huang2023guided}}& \makecell[c]{\textbf{1.00}\ci{.00}}& \makecell[c]{0.73\ci{.12}}& \makecell[c]{0.94\ci{.05}}& \makecell[c]{0.73\ci{.20}}& \makecell[c]{0.63\ci{.19}}& \makecell[c]{0.18\ci{.14}}& \makecell[c]{0.02\ci{.01}}& \makecell[c]{0.00\ci{.00}}& \makecell[c]{0.02\ci{.07}} & \makecell[c]{0.05\ci{.04}} \\ 
    \makecell[c]{\textbf{ViSkill}*\cite{huang2023value}}& \makecell[c]{-}& \makecell[c]{-}& \makecell[c]{-}& \makecell[c]{-}& \makecell[c]{-}& \makecell[c]{\textbf{0.85}\ci{.08}} & \makecell[c]{0.81\ci{.04}} & \makecell[c]{\textbf{0.57}\ci{.06}} & \makecell[c]{-}& \makecell[c]{-}\\ 
    \makecell[c]{\textbf{T-STAR}*\cite{lee2021adversarial}}&  \makecell[c]{-}& \makecell[c]{-}& \makecell[c]{-}& \makecell[c]{-}& \makecell[c]{-}& \makecell[c]{0.67\ci{.05}} & \makecell[c]{0.65\ci{.10}} & \makecell[c]{0.45\ci{.04}} & \makecell[c]{-}& \makecell[c]{-}\\ 
    \midrule
    \rowcolor{green!10} \textbf{Ours} & {\textbf{1.00}\ci{.00}}& {\textbf{0.95}\ci{.05}}& {\textbf{1.00}\ci{.00}}& {\textbf{0.84}\ci{.10}}& {\textbf{0.74}\ci{.12}}& {0.42\ci{.08}}& {\textbf{1.00}\ci{.00}}& {0.48\ci{.09}}& {\textbf{0.30}\ci{.12}}& {\textbf{0.37}\ci{.18}}\\
    \bottomrule

    \end{tabular*}
    \begin{tablenotes}    
    \scriptsize          
    \item ``*'' means the approach leverages manually pre-defined prior information for each specific task, such as task decomposition and chaining. ``-'' denotes not suitable.
    \end{tablenotes}
    \end{threeparttable}

    \label{comparison_results_table}
    \vspace{-2mm}
\end{table*}

\begin{figure*}[t]

    \centering
    
    \includegraphics[width=2.00\columnwidth]{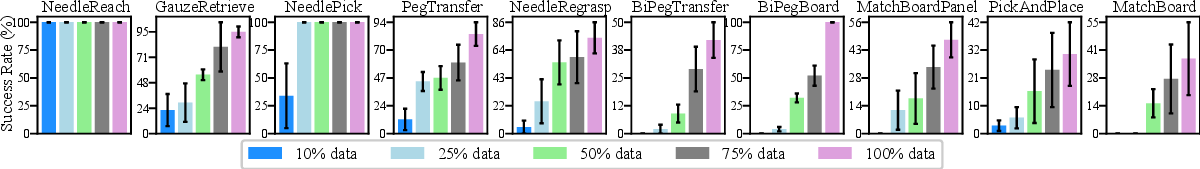}
    \setlength{\abovecaptionskip}{-0.15cm}

    \caption{Ablation results for the available data amount for the 10 different tasks. The mean value and standard deviation of the success rate for all tasks with different amount of available data are visualized to evaluate the performance of the trained model.}
    \label{ablation_study_fig}
    \vspace{-5mm}
\end{figure*}

In the experiments, we validate the performance of our proposed method in comparison with state-of-the-art methods using SurRoL simulator~\cite{long2023human}. We conduct ablation experiments on key variables of our model to quantify their effects. We finally show real-world dVRK deployment results.

\vspace{-2mm}
\subsection{Experiment setup}

\label{experiment_setup_subsection}
\textit{Tasks:} We consider skill-training surgical robotic tasks, which give goal states for the robots to reach and implement designed functions. To benchmark the performance of our approach, we adopt 10 state-based tasks as included in SurRoL ~\cite{surrol_web}: NeedleReach, GauzeRetrieve, NeedlePick, PegTransfer, NeedleRegrasp, BiPegTransfer, BiPegBoard, MatchBoardPanel, PickAndPlace, and MatchBoard. The corresponding dimensions of observation space $\mathbb{O}$, goal space $\mathbb{G}$, and action space $\mathbb{A}$ are described in the first row of Table \ref{comparison_results_table}. We refer readers to our project website~\cite{surrol_web} for more detailed descriptions and demos of all the tasks. For every task, we provide 100 trajectories as the original dataset, which are generated from an expert script with priori knowledge.

\textit{Evaluation metric:} 
In the evaluation process, we pretrain the policy on all the tasks with the cross-task objectives. For downstream task learning, we update the model on data of the specific task, only with the action prediction objective. When the model training was done, for testing of each task, we randomly set a start state (e.g., randomizing the initial positions of the needle and gripper for NeedlePick task) for 100 times, and count the number of successfully completions of the task. The success rate is the number of successful trials divided by the total number of trials.
To evaluate the stability of the model, we set 5 different seeds to generate the random starting states in experiments for each task, and report the average success rate and standard deviation of multiple runs.



\begin{table*}[t]
\setlength{\abovecaptionskip}{-0.15cm}
    \caption{Ablation study results for diverse pretraining configurations with mean and standard deviation for all tasks.}
    \centering
    \footnotesize
    \setlength{\tabcolsep}{0.94mm}
    \begin{threeparttable}
    \begin{tabular*}{2.0\columnwidth}{c|cccccccccc}
    \toprule
    \makecell[c]{Task} & \makecell[c]{\scriptsize {NeedleReach}} & \makecell[c]{\scriptsize {GauzeRetrieve}} & \makecell[c]{\scriptsize {NeedlePick}} & \makecell[c]{\scriptsize \text{PegTransfer}} & \makecell[c]{\scriptsize {NeedleRegrasp}} & \makecell[c]{\scriptsize {BiPegTransfer}} & \makecell[c]{\scriptsize {BiPegBoard}}& \makecell[c]{\scriptsize {MatchBoardPanel}}& \makecell[c]{\scriptsize {PickAndPlace}} & \makecell[c]{\scriptsize {MatchBoard}}  \\
    \midrule

    \makecell[c]{\textbf{w/o DP}\tnote{1}}& \makecell[c]{1.00\ci{.00}}& \makecell[c]{0.80\ci{.08}}& \makecell[c]{0.91\ci{.04}}& \makecell[c]{0.80\ci{.07}}& \makecell[c]{0.67\ci{.08}}& \makecell[c]{0.33\ci{.05}}& \makecell[c]{0.92\ci{.05}}& \makecell[c]{0.35\ci{.06}}& \makecell[c]{0.21\ci{.04}}& \makecell[c]{0.33\ci{.05}} \\  
    \makecell[c]{\textbf{w/o TP}\tnote{2}}& \makecell[c]{1.00\ci{.00}}& \makecell[c]{0.88\ci{.07}}& \makecell[c]{0.91\ci{.05}}& \makecell[c]{0.71\ci{.07}}& \makecell[c]{0.65\ci{.11}}& \makecell[c]{0.32\ci{.07}}& \makecell[c]{0.83\ci{.14}}& \makecell[c]{0.27\ci{.08}}& \makecell[c]{0.20\ci{.08}}& \makecell[c]{0.29\ci{.19}} \\ 
    \makecell[c]{\textbf{w/o SR}\tnote{3}}& \makecell[c]{1.00\ci{.01}}& \makecell[c]{0.92\ci{.04}}& \makecell[c]{0.93\ci{.01}}& \makecell[c]{0.72\ci{.08}}& \makecell[c]{0.67\ci{.09}}& \makecell[c]{0.31\ci{.15}}& \makecell[c]{0.91\ci{.07}}& \makecell[c]{0.36\ci{.04}}& \makecell[c]{0.26\ci{.07}}& \makecell[c]{0.25\ci{.06}} \\ 
    \makecell[c]{\textbf{w/o Pretrain}}& \makecell[c]{0.99\ci{.02}}& \makecell[c]{0.79\ci{.07}}& \makecell[c]{0.85\ci{.11}}& \makecell[c]{0.68\ci{.11}}& \makecell[c]{0.64\ci{.16}}& \makecell[c]{0.28\ci{.07}}& \makecell[c]{0.88\ci{.09}}& \makecell[c]{0.33\ci{.19}}& \makecell[c]{0.18\ci{.08}}& \makecell[c]{0.28\ci{.09}} \\
    \midrule
    \textbf{Ours} & {1.00\ci{.00}}& {0.95\ci{.05}}& {1.00\ci{.00}}& {0.84\ci{.10}}& {0.74\ci{.12}}& {0.42\ci{.08}}& {1.00\ci{.00}}& {0.48\ci{.09}}& {0.30\ci{.12}}& {0.37\ci{.18}}\\
    \bottomrule

    \end{tabular*}
    \begin{tablenotes}    
    \scriptsize
    \item $^{1}$ without dynamics prediction, $^{2}$ without time-to-goal prediction, $^{3}$ without sequence reconstruction.

    \end{tablenotes}
    \end{threeparttable}

    \label{ablation_results_table}
\end{table*}

\begin{figure*}[t]
\vspace{-4mm}
    \centering
    \setlength{\abovecaptionskip}{-0.3cm}
    \includegraphics[width=0.95\textwidth]{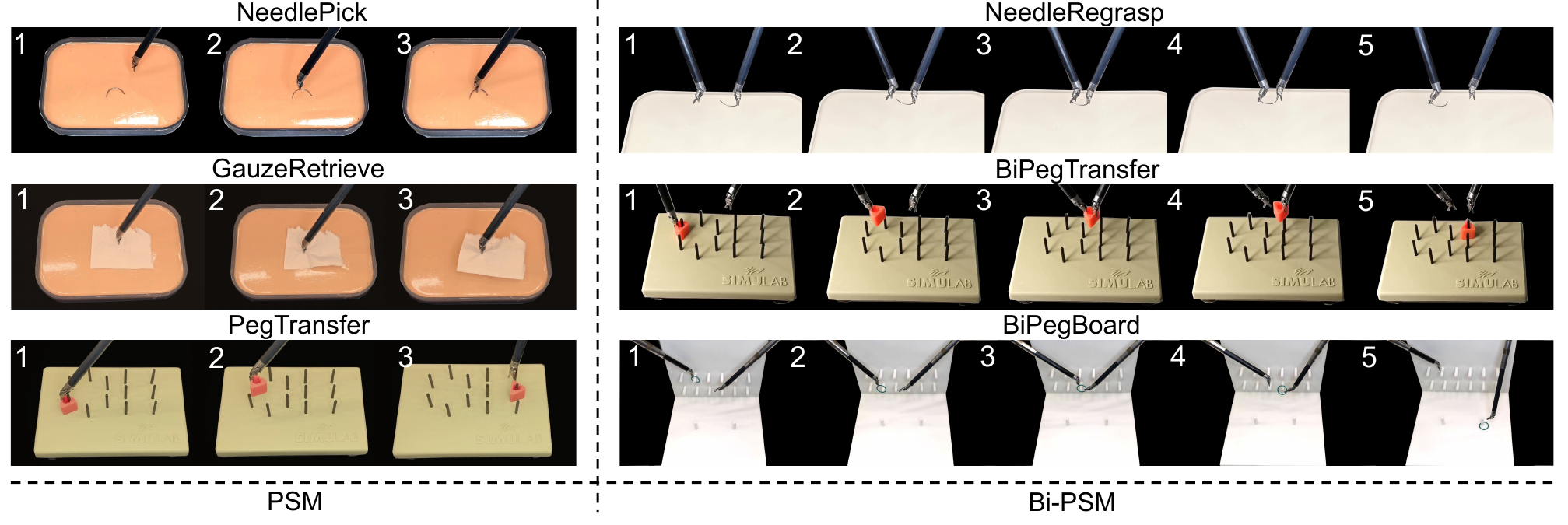}

    \caption{Illustration of the trajectory deployment of 6 tasks in dVRK platform. The timestep in the entire episode is labeled.}
    \vspace{-0.5cm}
    \label{dvrk_experiment_fig}
\end{figure*}

\vspace{-2mm}
\subsection{Comparison with state-of-the-art methods}

\label{comparison_experiment_subsection}

We provide both offline and online methods as the compared baselines. The offline methods are divided into two categories, offline reinforcement learning and imitation learning. For the offline reinforcement learning methods part, we involve \textbf{PLAS}~\cite{zhou2021plas} and \textbf{IQL}~\cite{kostrikov2021offline}. The offline reinforcement learning methods are implemented from d3rlpy library~\cite{d3rlpy} with original parameters. For the imitation learning methods part, we select \textbf{BC}~\cite{bain1995framework} and \textbf{VINN}~\cite{pari2021surprising}. The involved imitation learning methods are realized from our previous work~\cite{huang2023guided}. We also report the results from \textbf{DT}~\cite{NEURIPS2021_7f489f64}, which is a basic decision-making method based on modeling sequences with transformers. For each baseline, we deploy the proposed data augmentation method to maintain entirely fairness.

The main results are shown in Table \ref{comparison_results_table}. We observe that for the tasks with relatively low dimensions of state and action dimensions, e.g., NeedleReach, almost all the methods could get respectable success rates. For more complicated and long-horizon tasks, e.g., BiPegBoard and MatchBoardPanel, all the offline methods that are independent of the sequence reasoning capacity cannot reach the expected goal in the evaluations. Our model and \textbf{DT} ~\cite{NEURIPS2021_7f489f64} achieve success cases even in those sophisticated tasks with the sequence temporal modeling and reasoning ability of large transformer architecture from transformers. Compared with \textbf{DT}~\cite{NEURIPS2021_7f489f64}, our method shows an average 0.31 improvement on success rate since our design pretrain on multiple tasks with several extra losses and embeds the time-to-goal indicator. They guide future awareness and help learn the dynamics of the goal-conditioned paradigm to boost ultimate performance. 

Although our method does not need to interact with environment, we also compared with the state-of-the-art online methods, which consists of \textbf{DDPGBC}~\cite{nair2018overcoming}, \textbf{AMP}~\cite{peng2021amp}, \textbf{CoL}~\cite{goecks2019integrating}, \textbf{AWAC}~\cite{nair2020awac}, \textbf{DEX}~\cite{huang2023guided}, \textbf{ViSkill}~\cite{huang2023value}, and \textbf{T-STAR}~\cite{lee2021adversarial}. We implement those algorithms based on OpenAI library~\cite{henderson2018deep} and our previous work~\cite{huang2023guided,huang2023value}. We present the experimental results in Table \ref{comparison_results_table}. We find that compared with offline methods, the aggregate performance of the online methods has greatly improved since the active interaction with environments. However, powered by the strong reasoning among long sequences of the transformer architecture, our model still displays considerable performance compared to most of the online algorithms in involved tasks, only using 0.05 of required data in the training process. For those algorithms designed for specific tasks with pre-defined priori knowledge, such as \textbf{ViSkill}~\cite{huang2023value} and \textbf{T-STAR}~\cite{lee2021adversarial} in BiPegTransfer, BiPegBoard, and MatchBoardpanel, our model still surpasses or holds at almost equal levels in two of the tasks. However, those methods~\cite{huang2023value, lee2021adversarial} that depend on task-specific decomposition cannot be generalized to other tasks, which limits their valid scenarios and makes the overall performance of this type of algorithm inferior to ours.

\vspace{-2mm}
\subsection{Ablation study}

\label{ablation_study_subsection}

\subsubsection{Effect of cross-task pretraining}

The ablation results in different tasks of our cross-task pretraining strategy are shown in Table \ref{ablation_results_table}. We deploy the same model architecture but using different pretraining strategies. Our method improves the average success rate by 0.19 compared to methods that do not use pretraining. In addition, these models are not as robust as the model pretrained to other tasks and show a large variance in different test epochs, which shows that our cross-task pretraining improves stability through the understanding of the general task paradigm.

\subsubsection{Effect of multi-objective training}

We also provide the ablation results for each auxiliary component of the deployed training objectives in the pretraining process in Table \ref{ablation_results_table}. The evaluation results demonstrate that the involvement of each objective item can help boost the average success rate and limit the instability of the performance, which is attributed to reasoning capacity among sequence items and understanding of task dynamics of the transformer architecture, supported by the designed multiple training objectives.

\subsubsection{Influence of data amount}

In order to explore the impact of training data volume on experimental results, we split the original dataset of each task by the selected typical percentage. The multi-objective pretraining pattern among all the tasks and the proposed data augmentation approach is also deployed. From the results in Fig. \ref{ablation_study_fig}, we summarize that for the simple task, e.g., NeedleReach, even though the available data is constrained to a low amount, the final performance is still considerable. For other complicated tasks with richer state spaces and long horizons to complete, the performance collapses with the reduction of the data amount. This phenomenon means that more complex tasks require more data for training to ensure effectiveness of the model.

\vspace{-2mm}
\subsection{Real-world robot deployment}

\label{real_world_experiment_subsection}

The final trajectory of our model is deployed in dVRK equipment to evaluate its practicality and employability. The environmental scenes come from 3D printing of objects, which are the digital twin of SurRoL simulator~\cite{xu2021surrol,long2023human}. The involved tasks contain Needlepick, GauzeRetrieve, PegTransfer, NeedleRegrasp, BiPegTransfer, and BiPegBoard, where the first three tasks are with patient-sided manipulator (PSM) and the last three are with bimanual patient-sided manipulator (Bi-PSM). The finishing procedures of each task are shown in Fig. \ref{dvrk_experiment_fig}. The results illustrate that our method can support both PSM and Bi-PSM to manipulate the objects precisely, such as conveying pegs and regrasping needles. These findings illustrate that our model's predicted trajectories can be deployed in the real-world platform of dVRK,
thanks to digital twin design of SurRoL.


\vspace{-2mm}
\section{Conclusion}

We present goal-conditioned decision transformer, which leverages the large-scale transformers from LLMs to complete goal-conditioned surgical robot automation tasks. The proposed method depends on a novel sequence modeling with future indicators to realize temporal reasoning capability in goal-conditioned surgical tasks. Besides, we utilize multi-objective pretraining across multiple tasks to improve the sequence contextual reasoning and comprehension of the general goal-conditioned paradigm to anticipate precise decisions. Experiment results show that our method achieves superior performance and versatility across diverse tasks and the trajectory is deployable in dVRK to verify applicability.





\bibliographystyle{IEEEtran}

\bibliography{IEEEexample}

\end{document}